\documentclass[11pt]{article}

\usepackage[final]{acl}

\usepackage{times}
\usepackage{latexsym}

\usepackage[T1]{fontenc}

\usepackage[utf8]{inputenc}

\usepackage{microtype}

\usepackage{inconsolata}

\usepackage{graphicx}
\usepackage{amsmath,amssymb,amsfonts}
\usepackage{amsthm}
\usepackage{booktabs}
\usepackage{float} 
\usepackage{flushend}
\usepackage{graphicx}
\usepackage{mathrsfs}
\usepackage{multirow}
\usepackage{accents}
\usepackage{ragged2e}
\usepackage{setspace}
\usepackage{subfig}
\usepackage{tabularx}
\usepackage{textcomp}
\usepackage{url}
\usepackage{verbatim}
\usepackage{xcolor}
\usepackage{xspace}
\usepackage{tabularx}
\usepackage{svg}
\usepackage{balance}

\usepackage[noend]{algpseudocode}
\usepackage[linesnumbered,ruled,noend,vlined]{algorithm2e}

\usepackage{booktabs}
\usepackage{multirow}
\usepackage{siunitx}
\usepackage[T1]{fontenc}
\usepackage{inconsolata}
\usepackage{xcolor}
\definecolor{codebg}{RGB}{248,248,248}

\usepackage{minted}
\usepackage{listings}
\lstdefinestyle{mypython}{
  language=Python,
  backgroundcolor=\color{codebg},
  numbers=left,
  numberstyle=\tiny\color{gray},
  basicstyle=\ttfamily\small,
  keywordstyle=\color{blue},
  stringstyle=\color{magenta},
  commentstyle=\color{teal!70!black},
  showstringspaces=false,
  breaklines=true,
  tabsize=2,
  numbersep=6pt
}

\setminted{
  linenos,
  breaklines,
  autogobble,
  tabsize=2,
  fontsize=\small,
  numbersep=6pt,
  bgcolor=codebg
}

\usepackage{booktabs}
\usepackage{multirow}
\usepackage{siunitx}
\usepackage[most]{tcolorbox}

\usepackage{graphicx}

\usepackage{xcolor}
\usepackage{tikz}
\usetikzlibrary{positioning, shapes.geometric}

\usetikzlibrary{
    shapes.geometric, 
    positioning,      
    calc              
}

\usepackage{xcolor}
\usetikzlibrary{calc}

\definecolor{taxogray}{RGB}{242,242,242}
\usepackage{tikz}
\usepackage[edges]{forest}

\usepackage{hyperref}

\usepackage{booktabs}
\usepackage{array}
\usepackage{multirow}
\usepackage{graphicx}
\usepackage{hyperref}
\usepackage{amssymb}
\usepackage{booktabs}
\usepackage{array}
\usepackage{wasysym}

\newcommand{\fullcircle}{\CIRCLE}
\newcommand{\halfcircle}{\LEFTcircle}
\newcommand{\emptycircle}{\Circle}

\usepackage{multirow}
\usepackage{makecell}

\title{Are Large Language Models Suitable for Graph Computation?\\ Progress and Prospects}

\author{
  Yuting Zhang$^{1}$ \quad
  Yi Han$^{1}$ \quad
  Kai Wang$^{2}$ \quad
  Wei Ni$^{3}$ \quad
  Angela Bonifati$^{4}$ \quad
  Wenjie Zhang$^{1}$ \\
  $^{1}$University of New South Wales \\
  $^{2}$Antai College of Economics and Management, Shanghai Jiao Tong University \\
  $^{3}$Edith Cowan University
  $^{4}$Lyon 1 University\\
  \texttt{\{yutingz, zhangw\}@cse.unsw.edu.au} \quad
  \texttt{yi.han.1@student.unsw.edu.au} \\
  \texttt{w.kai@sjtu.edu.cn} \quad
  \texttt{w.ni@ecu.edu.au} \quad
  \texttt{angela.bonifati@univ-lyon1.fr} \quad
}

\begin{document}

\maketitle
\begin{abstract}
Large language models (LLMs) have been increasingly explored for graph computation, where tasks require reasoning over structured relationships and algorithmic operations. 
Yet, it remains unclear when LLMs can reliably support such computation and how they should be incorporated into graph-solving pipelines.  Existing surveys at the intersection of LLMs and graphs primarily focus on graph learning, text-attributed graphs, or graph-language modeling.
To bridge this gap, we provide a comprehensive review of LLMs for graph computation through a role-based taxonomy. 
Specifically, we identify two major paradigms: 
\romannumeral 1) LLMs as executors, where models directly solve graph tasks from graph descriptions and instructions; and \romannumeral 2) LLMs as planners, where models formulate problems, decompose reasoning steps, and invoke external tools or agents for execution. 
Based on this taxonomy, we analyze the strengths and limitations of current methods.
Our review indicates that LLMs are promising for simple, small-scale tasks, but remain unreliable for large-scale and exactness-demanding tasks.
Finally, we summarize available datasets and suggest four future directions.


\end{abstract}

\section{Introduction}
\label{sct:introduction}

Graphs provide a natural representation for relationships in many real-world systems, such as social connections \cite{breuer2020friend}, protein-protein interactions \cite{neyshabur2013netal}, and paper citations \cite{gollapalli2014extracting}. 
Analyzing such graph-structured data often requires graph computation, which derives explicit structural results or quantities from a given graph, such as shortest paths \cite{dijkstra2022note}, triangle counts \cite{azad2015parallel}, and cycle findings \cite{qiu2018real}. Such computation underlies a wide range of applications, including
social network analysis \cite{newman2003structure}, 
fraud detection \cite{lyu2020maximum},
and recommendation systems \cite{he2017neural}.

However, the reliance on specialized knowledge of graph computation techniques creates a significant barrier to applying graph-based analytics to real-world problems. Therefore, there is a need for approaches that can leverage LLMs to interpret user intent and translate it into computational solutions, lowering the barrier to graph computation \cite{liu2024one}.
Despite the needs and growing attention to the use of LLMs in addressing graph computational tasks \cite{guo2025g1}, whether LLMs are fundamentally suitable for graph computation remains controversial.
Graph computation often requires exact structural reasoning and faithful tracking of vertices and edges or multi-step algorithmic execution, which differ substantially from the probabilistic next-token prediction objective of LLMs.
Existing studies have shown that LLMs may hallucinate graph structures, miss critical edges, and suffer from severe performance degradation as graph size or reasoning depth increases \cite{liu2024lost, wang2024microstructures, heyman2025reasoning}.
These findings raise concerns about using LLMs for graph computation, especially in correctness-critical scenarios.




As compared in Table \ref{tab:survey_comparison}, although a limited number of surveys have explored LLMs for graphs, none have systematically examined the suitability of LLMs for graph computation.
\citet{ren2024survey} review methods that integrate LLMs with graph learning techniques, while \citet{li2023survey} focus on text-attributed graphs and categorize methods by the roles of LLMs in the model pipeline.
\citet{jin2024large} survey various graph types, but do not clearly distinguish graph computation from graph learning for graphs with limited textual semantics.
Focusing on retrieval-augmented generation with graphs, \citet{han2025graphrag} briefly mention graph computation when discussing LLM-based graph reasoning.
Therefore, despite the rapid growth of this area, the literature on LLMs for graph computation remains insufficiently organized.


\begin{table}[t]
\centering
\scriptsize
\setlength{\tabcolsep}{3pt}
\renewcommand{\arraystretch}{1.1}
\caption{Comparison of existing surveys on LLMs for graph from four perspectives:
graph-computation focus, task specification, performance comparison on reported experimental results, and dataset collection.}
\label{tab:survey_comparison}

\begin{tabular*}{\columnwidth}{@{\extracolsep{\fill}}lcccc@{}}
\toprule
\textbf{Survey}
& \shortstack{\textbf{Graph-}\\\textbf{comp. focus}}
& \shortstack{\textbf{Tasks}\\\textbf{specification}}
& \shortstack{\textbf{Performance}\\\textbf{summarization}}
& \shortstack{\textbf{Dataset}\\\textbf{collection}} \\
\midrule

\cite{ren2024survey}
& \halfcircle & \emptycircle & \emptycircle & \emptycircle \\

\cite{jin2024large}
& \halfcircle & \fullcircle & \emptycircle & \fullcircle \\

\cite{li2023survey}
& \halfcircle & \emptycircle & \emptycircle & \emptycircle \\

\cite{han2025graphrag}
& \halfcircle & \halfcircle & \emptycircle & \fullcircle \\

Ours    
& \fullcircle & \fullcircle & \fullcircle & \fullcircle \\

\bottomrule
\end{tabular*}

\vspace{6pt}
\begin{minipage}{\columnwidth}
\footnotesize
\fullcircle: central and explicit coverage;
\halfcircle: explicit but partial coverage;
\emptycircle: no dedicated coverage.
\end{minipage}
\end{table}

This survey aims to bridge this gap by providing an in-depth and systematic review focusing on LLMs for graph computations and exploring potential avenues for future research. As shown in Table \ref{tab:survey_comparison}, our survey focuses on LLMs for graph computation and explicitly provides definitions of related tasks, performance comparisons based on reported experimental results in each paper, and dataset collections within the scope. The contributions can be summarized as follows:

\noindent {\em $\bullet$ } \textit{Role-based taxonomy.}
We introduce a new taxonomy for categorizing research in LLMs for graph computational tasks. 
Our taxonomy clarifies the roles that LLMs participate in the graph computational tasks.
Building upon this foundation, we further sub-categorize the research based on the specific techniques employed within each paradigm.


\noindent {\em $\bullet$ } \textit{Comprehensive review and summary.} Based on the proposed taxonomy, we review the current literature, analyzing the strengths and limitations of each method. We also summarize the definitions and time complexity of representative graph computational tasks in Appendix \ref{sec:graph_tasks}. In Appendix \ref{sec:benchmark}, we assemble an extensive set of datasets for evaluating LLMs on graph computation. We also compare representative methods with their reported performance and code availability in Appendix \ref{sec:methods}.

\noindent {\em $\bullet$ } \textit{Evidence-based assessment and future directions.} Based on existing evidence, we argue that LLMs are promising as flexible interfaces for easy tasks and small graphs, but are limited for large-scale graphs and correctness-critical graph tasks.
Based on this analysis, we further propose four future directions, including developing semantic graph benchmarks, optimizing multi-step execution for complex graph queries, preserving the privacy of graphs in prompts and training data, and adapting LLMs to domain-specific graph structures.


\tikzset{
    rootnode/.style={draw=blue!50, rounded corners=3pt, fill=blue!8, line width=0.7pt, inner sep=4pt, align=center, font=\normalsize\bfseries},
    category/.style={draw=teal!50, rounded corners=3pt, fill=teal!8, line width=0.7pt, inner sep=4pt, align=center, font=\small\bfseries},
    leafnode/.style={draw=orange!50, rounded corners=3pt, fill=orange!8, line width=0.7pt, inner xsep=3pt, inner ysep=3pt, align=center, font=\small, text width=1.6cm},
    citebox/.style={draw=gray!50, rounded corners=3pt, fill=gray!4, minimum width=2.2cm, minimum height=0.5cm, inner xsep=3pt, inner ysep=3pt, text=gray!100!black, font=\footnotesize, text width=11.5cm, align=flush left}
}

\begin{figure*}[ht!]
\centering

\begin{forest}
  for tree={
    grow'=0,
    forked edges,
    s sep=2.1mm,
    l sep=2.2mm,
    anchor=west,
    child anchor=west,
    parent anchor=east,
    edge={thick, -stealth, draw=gray!70, rounded corners=0pt},
  }
  [{\rotatebox{90}{Taxonomy}}, rootnode
    [\rotatebox{90}{LLMs as Executors}, category
      [Prompting, leafnode
        [{NLGraph~\cite{wang2023can}, GPT4Graph~\cite{guo2023gpt4graph}, LLMtoGraph~\cite{liu2023evaluating}, LLM4DyG~\cite{zhang2024llm4dyg}, GraCoRe~\cite{yuan2025gracore}, GraphArena~\cite{tang2024grapharena}, GraphPattern~\cite{dai2025large}, LLMTM \cite{hao2026llmtm}}, citebox]
      ]
      [Encoding, leafnode
        [{Talk Like a Graph~\cite{fatemi2023talk}, GraphLLM~\cite{chai2025graphllm}, Graph-Token~\cite{perozzi2024let}, GraphDO~\cite{ge2025can}, GraphOmni~\cite{xu2025graphomni}, GraphInsight \cite{cao2025graphinsight}, \cite{xypolopoulos2024graph}, CL-OWL~\cite{zangari2026colorful}, DynamicTRF \cite{wei2025harnessing}, DynamicGTR \cite{wei2026dynamicgtr}
        }, citebox]
      ]
      [Post-training, leafnode
        [{GraphWiz~\cite{chen2024graphwiz}, GITA~\cite{wei2024gita}, InstructGraph~\cite{wang2024instructgraph}, GraphInstruct~\cite{luo2024graphinstruct}, NLGift~\cite{zhang2024can}, HLM-G~\cite{khurana2024hierarchical}, GraphPRM~\cite{peng2025rewarding}, G1~\cite{guo2025g1}, \cite{zhang2025generalizable}, GraphThought \cite{huang2025graphthought}, \cite{shirai2025less}, NPG-Muse \cite{wang2025graph}}, citebox]
      ]
    ]
    [\rotatebox{90}{LLMs as Planners}, category
      [Code Generation, leafnode
        [{GraphArena~\cite{tang2024grapharena}, GCoder~\cite{zhang2024gcoder}, GRRAF \cite{li2025grraf}, PIE~\cite{gong2025pseudocode}, GraphMind~\cite{zhang2025improving}, Simple-RTC~\cite{hu2025rethinking}, GraphSkill \cite{wang2026graphskill}}, citebox]
      ]
      [Function Calling, leafnode
        [{Graph-ToolFormer~\cite{zhang2023graph}, LLM4Graph~\cite{li2024can}, Graph-CoT \cite{jin2024graphcot}, Graph-Grounded LLMs~\cite{gupta2025graph}, GraphChain~\cite{wei2026graphchain}, GraphTool-Instruction~\cite{wang2025graphtool}, EGL-SCA \cite{yuan2026eglsca}}, citebox]
      ]
      [Multi-agent \\Collaboration, leafnode
        [{GraphTeam~\cite{li2024graphteam}, GraphAgent-Reasoner~\cite{hu2024scalable}, MA-GTS~\cite{yuan2025ma}, GTA~\cite{xu2025gta}, GraphDC\cite{li2026graphdc}, GraphVista\cite{han2026graphvista}, GraphCogent\cite{wang2026graphcogent} }, citebox]
      ]
    ]
  ]
\end{forest}

\caption{Taxonomy of LLMs for graph computation.}
\label{fig:taxonomy}
\end{figure*}
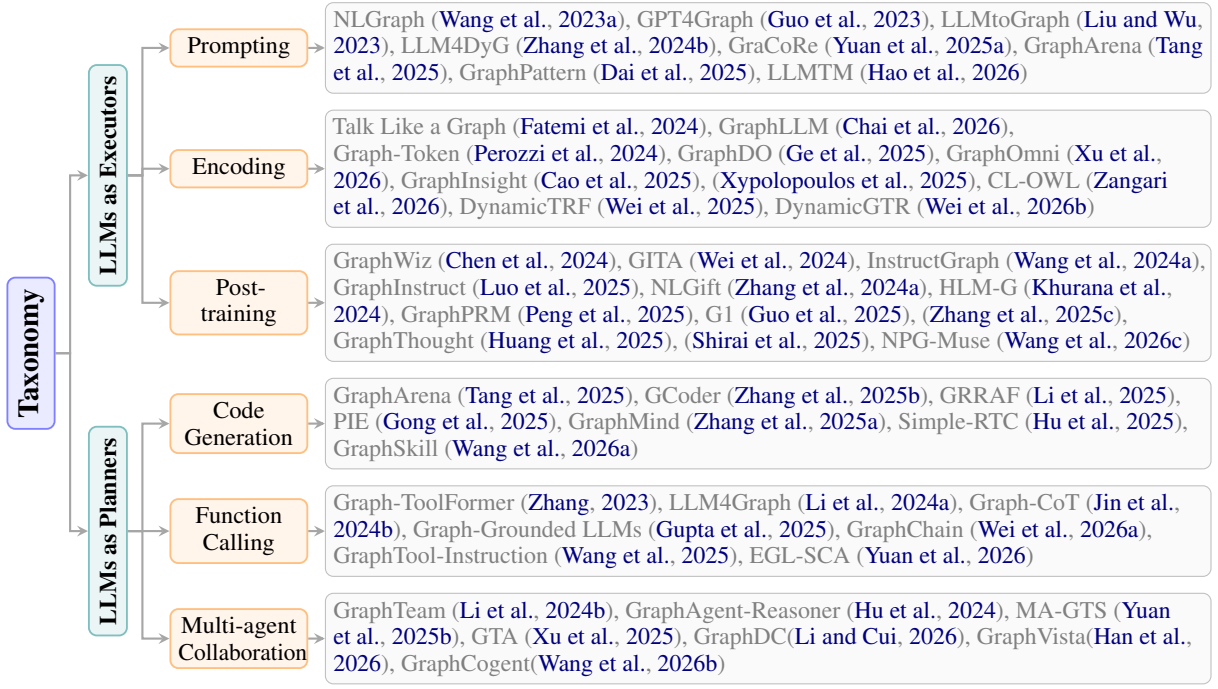

\section{Background}
\label{sec:background}
In this section, we first provide essential background knowledge on graph computational tasks and LLMs. Then, we present our taxonomy of LLMs for graph computation.

\subsection{Definitions}
\noindent
\textbf{Graph-structured data.}
A graph $G = (V, E)$ is a non-linear data structure that consists of a set of vertices $V$, and a set of edges $E$ connecting these vertices. 
Each edge $e \in E$ is associated with a pair of vertices $(u,v)$, where $u$ and $v$ are the endpoints of the edge. 
Depending on the specific information carried by their vertices and edges, graphs can be classified into various types.
For example, directed graphs have directional edges $e =(u,v)$ pointing from $u$ to $v$, and edge-weighted graphs have a weight $w(e) \in \mathbb{R}$ for each edge.

\noindent
\textbf{Graph description.}
As LLMs rely on a sequential text interface, describing the graph structure textually is the fundamental step for enabling LLM-based graph computation.
List structures \cite{tabassum2018social}, such as edge list and adjacency list, are commonly considered as graph representation methods due to their space efficiency, which helps minimize token consumption when encoding graph data for LLMs.

\noindent {\em $\bullet$ } The adjacency list $\mathcal{L}$ of a graph $G$ with $n$ vertices is a collection of $n$ lists, where the list $l_u$ for each vertex $u$ contains all vertices $v$ such that there is an edge from $u$ to $v$, i.e., $l_u=\{v \mid (u,v) \in E\}$. 

\noindent {\em $\bullet$ } The edge list $\mathcal{E}$ of a graph $G$ is a list of all edges in the graph. Each edge is represented as a pair $(u, v)$, where $u$ and $v$ are the endpoints of the edge, i.e., $\mathcal{E} = \{\, (u, v) \mid (u, v) \in E \,\}$.

Besides list-based representations, matrix-based representations, such as adjacency matrix \cite{harary1962determinant} and incidence matrix \cite{fulkerson1965incidence}, and graph description languages such as Graph Modeling Language \cite{himsolt1997gml}, and Graph Markup Language \cite{brandes2010graph}  are also widely considered to describe graphs.






\noindent
\textbf{Graph computational tasks.} Graph computational tasks are problems defined on graphs with the objective of discovering relational patterns, extracting structural properties, or computing mathematical metrics of a given graph. 
Unlike graph learning tasks, e.g., node classification and link prediction, which produce learned predictions by learning predictive models or latent representations from graph data, graph computational tasks are solved through algorithmic operations on graph structures and produce deterministic results.
Formally, given a graph $G=(V, E)$, a graph computation task $\mathcal{T}$ refers to a problem that requires operations on the vertices and edges of $G$ to produce the answer $\mathcal{A}$. 
We have summarized the common tasks in Appendix \ref{sec:graph_tasks}.





\noindent
\textbf{Large language models.}
LLMs are highly parameterized (i.e., billion-level) language models that are pre-trained on massive text corpora and designed to understand and generate content from text distributions \cite{zhao2023survey}. 
LLMs can be easily and efficiently fine-tuned on task-specific data to achieve better results across diverse downstream tasks. 
Beyond standard statistical text generation, LLMs can act as the central reasoning engine and be augmented with dynamic memory and external computational tools to function as autonomous agents. 
By giving the model access to external tools and memory, LLMs have demonstrated impressive capabilities in planning multi-step strategies and solving complex questions \cite{zhao2024expel, mohammadi2025evaluation}.


\noindent
\textbf{LLMs for graph computation.} LLMs for graph computation refer to methods that involve LLMs in the process of solving graph computational tasks. 
Given the textual description of graph $G = (V, E)$ and a graph computation task $\mathcal{T}$ as the input, LLMs leverage their pre-trained knowledge, reasoning capabilities, or the provided external tools to produce the corresponding answer $\mathcal{A}$.

\subsection{Taxonomy}
In this survey, we organize existing works by the role of LLMs in graph computation.
As shown in Fig. \ref{fig:taxonomy}, we summarize two main paradigms: LLMs as Executors and LLMs as Planners.


\noindent {\em $\bullet$ } \textbf{LLMs as executors.} LLMs directly leverage their inherent reasoning abilities to solve graph problems. Existing executor-based methods mainly include three types: \romannumeral 1) \textit{Prompting}, which uses various prompt designs to describe tasks;
\romannumeral 2) \textit{Encoding}, which focuses on representing graph structures to LLMs via natural language or specialized encoders; and \romannumeral 3) \textit{Post-training}, which adapts LLMs to graph computation through fine-tuning on domain-specific graph datasets.

\noindent {\em $\bullet$ } \textbf{LLMs as planners.} LLMs serve as high-level orchestrators for analyzing and solving graph computational tasks. This includes three primary approaches: \romannumeral 1) \textit{Code generation}, where LLMs generate executable code to solve graph computational tasks; \romannumeral 2) \textit{Function calling}, where LLMs invoke predefined graph APIs or specialized tools to extract a result; and \romannumeral 3) \textit{Multi-agent collaboration}, where multiple LLM-based agents cooperate to solve graph computational tasks.

In the following sections, we review representative studies under each category and discuss their strengths and limitations.



\section{LLMs as Executors}
\label{sec:executor}

In this section, we focus on the paradigm of LLMs as executors. In this setting, LLMs are given a graph structure and a task instruction, perform reasoning internally, and directly output the final answer without relying on external tools. We comprehensively review the subcategories and discuss their advantages and limitations. 



\vspace{-1.5mm}
\subsection{Prompting} 

To equip LLMs for graph computation without modifying their internal parameters, a robust line of research investigates advanced prompting techniques. This approach leverages the models' inherent reasoning capabilities and pre-trained knowledge to interpret graph tasks strictly through natural language instructions.

Early studies investigate how to adapt LLMs to graph computation.
GPT4Graph \cite{guo2023gpt4graph} is among the first works to introduce a pipeline to integrate LLMs into graph computation.
Beyond general prompting strategies such as zero-shot prompting \cite{kojima2022large}, few-shot prompting \cite{brown2020language}, chain-of-thought \cite{wei2022chain}, least to most prompting \cite{zhou2022least}, and self-consistency \cite{wang2022self}, \citet{wang2023can} further enhance LLMs' capability on graph computation through Build-a-Graph Prompting and Algorithmic Prompting.


Recent studies have improved prompting methods to enable LLMs to handle a broader range of graph types and tasks. LLM4DyG \cite{zhang2024llm4dyg} and LLMTM \cite{hao2026llmtm} evaluate dynamic graphs where each edge is associated with a timestamp $t$ to capture the temporal evolution of a network. GraCoRe \cite{yuan2025gracore} evaluates heterogeneous graphs that include rich, multi-typed semantic information.
To push the boundaries of task diversity, investigations have explored whether LLMs can reliably enumerate all valid solutions to multi-answer queries, such as finding all shortest paths \cite{liu2023evaluating}. GraphArena \cite{tang2024grapharena} examines LLMs on solving polynomial-time tasks alongside NP-complete tasks on real-world datasets.
Furthermore, the work \cite{dai2025large} focuses on the tasks related to mining graph patterns, such as triangle, square, diamond, and house.



\subsection{Encoding}
Encoding is a crucial factor that can significantly affect LLMs' performance on graph computation, as it serves as an important preprocessing technique for representing graph structures in sequence form.

A major line of work improves LLM performance by designing more effective textual representations of graphs. 
GraphQA \cite{fatemi2023talk} experiments with different graph representations.
GraphDO \cite{ge2025can} suggests that the optimal ordering strategy is closely tied to the inherent structural requirements of the target task. 
To help LLMs better leverage the graph structures, GraphInsight \cite{cao2025graphinsight} places the most critical subgraph descriptions at the beginning and the end of the text sequence. \citet{xypolopoulos2024graph} propose graph linearization methods that renumber vertices according to structural properties such as centrality and degeneracy, and CL-OWL \cite{zangari2026colorful} introduces a human-interpretable structural encoding scheme that uses colors to represent graphs.
Besides, GraphOmni \cite{xu2025graphomni} proposes an RL-based adaptive mechanism to select the optimal combination of graph description and prompt. 

Beyond purely textual graph representations, another line of work injects graph structural knowledge into LLMs through specialized encoders. 
GraphLLM \cite{chai2025graphllm} employs a graph transformer to encode the graph structure into continuous vector embeddings, which are then used as prefix embeddings for LLMs.
Graph-Token \cite{perozzi2024let} explores various GNN architectures to convert the graph into soft tokens that are fed into the LLM. 
Beyond embedding-based fusion, 
DynamicTRF \cite{wei2025harnessing} and DynamicGTR \cite{wei2026dynamicgtr}
propose frameworks that dynamically select the most suitable textual or visual graph representation, optimizing both answer accuracy and response efficiency.


\subsection{Post-training}
Post-training methods adapt pre-trained LLMs for graph computation by optimizing them on graph-specific data. Compared with prompting and encoding, which keep model parameters fixed, post-training modifies the model to improve task-specific accuracy \cite{sanford2024understanding}.



To improve the generalization capabilities of the fine-tuned model, some works adopt instruction tuning. GraphInstruct \cite{luo2024graphinstruct} introduces a comprehensive benchmark covering 21 diverse tasks and introduces a label mask training to help LLMs learn node ID information during fine-tuning.
NLGift \cite{zhang2024can} shows that fine-tuned LLMs may still rely on memorized graph patterns rather than general graph reasoning rules.
To improve reasoning supervision, GraphThought \cite{huang2025graphthought} generates training trajectories through heuristic-guided forward reasoning and solver-guided backward reasoning.
Less is More \cite{shirai2025less} further shows that LLMs instruction-tuned on textual input can outperform models with specialized graph encoders.

Several works improve reliability and reduce hallucinations via Direct Preference Optimization (DPO), which aligns models with preference pairs, and Reinforcement Learning (RL), which optimizes via reward feedback.
By using DPO, GraphWiz \cite{chen2024graphwiz} 
further enhances the performance of instruction-tuned LLMs to align the model with the correct reasoning path.
To build high-quality preference pairs for DPO, InstructGraph \cite{wang2024instructgraph} constructs hallucination-based negative examples, and GraphPRM \cite{peng2025rewarding} uses Process Reward Models to score intermediate reasoning steps.
\citet{zhang2025generalizable} show that post-training alignment on synthetic graph data can partially transfer to real-world tasks with implicit graph structures.
By adopting RL, G1 \cite{guo2025g1} enables LLMs to self-improve their graph reasoning abilities through rule-based verifiable rewards,
and NPG-Muse \cite{wang2025graph} introduces NP-hard graph problems as a training corpus.

To enable LLMs on large graphs, several studies have explored methods to reduce the computational cost of the prompts, which can become lengthy when incorporating both feature and structural information from graphs.
HLM-G \cite{khurana2024hierarchical} uses bespoke attention masks to force the token processing to remain localized within descriptions of single vertices, reducing the computational complexity from $\mathcal{O}\left((\sum{n_i})^2\right)$ to $\mathcal{O}\left(\sum{n_i}^2\right)$, where $n_i$ is the number of tokens describing vertex $v_i$.
GITA \cite{wei2024gita} integrates fine-tuned vision-language models into LLM-based graph computation and adopts $k$-hop subgraphs sampling, where the subgraph for vertex $u$ is induced by all vertices within $k$ steps from $u$.

\subsection{Summary}
Methods reviewed in this section improve the ability of LLMs to directly execute graph computation. 
Prompting-based methods improve model performance by encouraging models to think or follow intermediate computational steps. Encoding-based methods improve accuracy by presenting graph structures in forms that are easier for LLMs to parse and reason over. Post-training-based methods improve accuracy by directly optimizing LLMs on graph computation data. Compared with prompting and encoding, post-training can generally produce better task-specific performance.

Although all three lines of work aim to make LLMs better graph executors, each category has its own limitations. For prompting, designing effective prompts, such as few-shot and CoT, requires human effort and graph algorithm knowledge. Different tasks also require task-specific prompts, limiting the flexibility of prompting methods across diverse graph tasks. For encoding, when the graph description becomes too long, models tend to forget the middle section \cite{liu2024lost, hsieh2024ruler}. Although placing the critical subgraph at the head and tail can improve the performance, it introduces additional cost to design the order. 
Post-training methods are restricted by their dependence on high-quality training datasets. Constructing such datasets is costly, as they typically necessitate broad coverage over diverse graph tasks and graph types. Moreover, LLMs often exhibit performance degradation on tasks or graphs that are not in the training data \cite{chen2024graphwiz}.
\section{LLMs as Planners}
\label{sec:planner}

Although using LLMs as executors offers convenience and accessibility for general users, LLMs are probabilistic sequence models, which are fragile to operate like deterministic symbolic executors with guaranteed mathematical or algorithmic correctness \cite{hahn2020theoretical,dziri2023faith}.
The planner paradigm shifts the LLMs from direct graph solving to external support such as programs, tools, or specialized agents, so that hallucination, omitted steps, and contextual overflow can be effectively avoided as the graph size and reasoning depth increase \cite{liu2024lost,heyman2025reasoning,wang2024microstructures,han2026graphvista}.
We review this paradigm through code generation, function calling, and multi-agent collaboration.

\subsection{Code Generation}

LLMs have demonstrated strong natural-language-to-code generation ability, which shifts the uncertainty of token-level reasoning into a more reliable computational process \cite{gao2023pal,chen2022program,chen2021evaluating,li2022competition}. 


GraphArena \cite{tang2024grapharena} first mentions code writing for graph computation and shows that executable solving can reduce errors caused by the repetitive iterative and backtracking operations required by many graph algorithms. 
Following this direction, GCoder \cite{zhang2024gcoder} collects code-oriented instruction-tuning data for graph computational problems and improves the LLM's graph-solving ability through supervised fine-tuning on code generation. GraphMind \cite{zhang2025improving} introduces a dataset featuring code-based solutions and execution traces, guiding LLMs to tackle graph problems programmatically. Simple-RTC \cite{hu2025rethinking} separates problem formatting, data extraction, algorithm reasoning, and code generation into a pipeline. 

Beyond simply generating code, several studies explore how LLMs can detect and handle execution errors during graph computation.
PIE \cite{gong2025pseudocode} injects task-related pseudocode into prompts and uses repeated trial-and-error execution, improving accuracy while keeping computation cost controlled. 
GRRAF \cite{li2025grraf} provides a training-free variant by storing the graph in a graph database and prompting the LLM to generate executable NetworkX or Cypher queries, with error feedback and timeout control to keep graph access outside the model context. 
GraphSkill \cite{wang2026graphskill} extends this code-based direction by combining hierarchical retrieval over technical documentation with self-generated small test cases, allowing the coding agent to reduce noisy retrieval and address logical errors.  

\subsection{Function Calling}
Function calling provides a more stable and constrained planner interface for graph computation, by allowing LLMs to select predefined graph APIs or tools from pre-existing graph toolkits or user-built libraries \cite{schick2023toolformer,patil2024gorilla,li2023apibank,patil2025bfcl}.  


Some studies explore how LLMs can invoke predefined graph tools. For instance, 
Graph-ToolFormer \cite{zhang2023graph} teaches the model to insert graph loading and graph analysis calls into generated statements.
LLM4Graph \cite{li2024can} proposes a pipeline based on API documents of frequently-used graph libraries, enabling LLMs to leverage both document and code to analyze larger-scale graphs. 
GraphTool-Instruction \cite{wang2025graphtool} provides a more effective tool-using pipeline by explicitly decomposing graph reasoning into graph extraction, tool-name identification, and tool-parameter extraction.

Recent work extends function calling to iterative settings.
Graph-CoT \cite{jin2024graphcot} introduces an iterative reasoning--interaction--execution framework for stepwise graph computation, allowing LLMs to focus on the information required at each step.
Graph-Grounded LLMs \cite{gupta2025graph} connect LLMs with a graph library through closed-loop function calls
to reduce hallucinations and mathematical inaccuracies. 
GraphChain \cite{wei2026graphchain} enables large-scale graph computation by letting LLMs compose graph tool-chains and compresses task-relevant graph information step by step.
EGL-SCA \cite{yuan2026eglsca} improves graph tool-using agents by employing structural credit assignment, which routes failures to either instruction updates or tool repairs.

\subsection{Multi-agent Collaboration}

Multi-agent collaboration provides a robust solution for complex graph problems by distributing graph-solving responsibilities across specialized agents. In this way, it alleviates the single-planner brittleness of the long reasoning step error accumulation as the graph scale and difficulty increase.




Some works follow a conventional multi-agent workflow design by assigning each agent a role in the graph-solving pipeline. 
GraphTeam \cite{li2024graphteam} organizes graph solving into cooperating agents with different specialties, converting graph analysis from one-shot reasoning into a staged process.
MA-GTS \cite{yuan2025ma} follows a similar role-specialized logic and extends to real-world graph computation problems where graph data may be implicit, noisy, and irregular.

To further improve the scalability, some recent works decompose the graphs across multiple agents. GraphAgent-Reasoner \cite{hu2024scalable} allocates a dedicated agent to process the information of a single vertex, employing a master LLM to organize the distributed computation and merge the answers. However, this method introduces computation overhead in distributed computations. 
GraphDC \cite{li2026graphdc} optimizes the computational cost by decomposing the graph into subgraphs rather than individual vertices, and letting the master agent process inter-subgraph edges.
To address memory and scalability constraints, GraphCogent \cite{wang2026graphcogent} employs a sensory agent to partition large complex graphs into manageable overlapping subgraphs via a sliding window, which are subsequently merged and indexed by a buffer agent.

Besides, some works decompose complex graph tasks into sub-tasks.
GTA \cite{xu2025gta} introduces a fine-tuned algorithm generator and algorithm decomposer to break down the graph tasks into manageable sub-steps.
GraphVista \cite{han2026graphvista} introduces a planning agent that decomposes graph tasks and adaptively assigns them to textual retrieval or visual reasoning modules based on task requirements and structural dependencies.

\subsection{Summary}

Methods reviewed in this section improve graph computation by shifting LLMs from direct executors to planners that coordinate external computational resources. 
Code generation provides the most flexible planner interface, allowing LLMs to translate graph problems into executable programs and substantially improve accuracy and scalability by delegating graph-state tracking and algorithm execution to interpreters.
Function calling enables LLMs to invoke predefined external tools, allowing complex graph operations to be performed more stably and efficiently, especially when the task matches an existing tool schema.
By assigning different agents to specialized roles, multi-agent collaboration has the flexibility of code generation with the reliability of function calling, enabling the planner to handle more complex graph problems.

However, each category also introduces distinct limitations. Code generation is prone to syntax errors, runtime bugs, API misuse, and input-format mismatches during the translation from a graph problem to executable code. 
Function calling is more constrained by tool coverage and compositionality. When it encounters an out-of-distribution operation, it may force the problem into the closest available function, causing invalid calls.
Multi-agent collaboration suffers from computational overhead compared to the previous methods, as it requires intermediate message exchanges across multiple interaction turns \cite{tran2026single}. Also, the intermediate errors may propagate across agents, leading to repeated and invalid communication, latency, and tool invocation.

\section{Discussions and Future Directions}
\label{sec:future}


In this section, we discuss whether LLMs are suitable for graph computation tasks based on existing literature and outline promising future directions.

\subsection{Discussions}
LLMs as executors offer an effective, convenient, and accessible method for ordinary users to solve a wide range of graph problems simply by providing a textual input. However, this direct execution approach is constrained to smaller graphs and only yields high accuracy on relatively simple tasks.
By shifting exact graph computation from the LLMs to external tools,
planner-based approaches achieved near-perfect performance on several widely used benchmarks 
and demonstrate strong scalability and effectiveness on large-scale graphs.
Overall, LLMs are suitable for assisting graph computation effectively as planners that interpret, decompose, and delegate tasks to reliable tools or code, while as standalone executors, they are suitable only for small graphs and for scenarios where accuracy requirements are not stringent.

Despite these promising capabilities, several challenges remain that limit a more reliable adoption of LLMs for graph computation.




\noindent
\textit{Scalability.} 
Most executor-based methods achieve reasonable performance only on small graphs of up to 50 nodes \cite{cao2025graphinsight}.
As the performance often drops when relevant information appears in the middle of a long context \cite{hu2024scalable, liu2024lost}, LLMs struggle to faithfully memorize the graph structures when textual descriptions become longer.
Furthermore, the limited context windows of LLMs prevent them from processing graphs whose descriptions exceed the maximum context length.
Planner-based systems suffer from context and inference overhead from operations like compiler feedback, API documents, compressed graph states, or inter-agent communication \cite{li2024can,gong2025pseudocode,wei2026graphchain,hu2024scalable}.
These operations consume considerable tokens and computation time, limiting the scalability to larger graphs.


\noindent
\textit{Generalization}. 
Executor-based methods tend to memorize graph patterns and training tasks rather than acquiring generalizable graph reasoning principles \cite{zhang2025generalizable}. Their performance may degrade significantly when evaluated on unseen graphs and tasks, indicating a reliance on superficial input patterns over true algorithmic understanding \cite{zhang2024can}.
Additionally, LLM performance is highly sensitive to text and structural representations. Different graph tasks and structures are better suited to specific prompting and encoding methods, necessitating careful optimization of these inputs \cite{fatemi2023talk, xu2025graphomni}.
Some planner-based methods heavily depend on unified graph description methods, task formulations, tool interfaces, and execution environments. 
Such requirements limit their generalization ability in real-world scenarios, where tasks, available tools, and environment constraints may vary significantly across domains.

\subsection{Future Directions}
Apart from the common objectives of expanding graph scale, tackling harder tasks, improving accuracy, and minimizing token consumption, several other promising directions remain unexplored.


\noindent {\em $\bullet$} 
\textit{Benchmarks for queries with implicit graph structures.}
Current benchmarks for LLM-based graph computation mostly provide explicit graph structures. In many real-world scenarios, graphs are often implicit in queries or documents, such as transaction logs and medical records \cite{wang2024microstructures, hu2023beyond}. In such cases, LLMs need to recognize and identify the underlying graph structure \cite{hsieh2024ruler,bai2025longbenchv2}. Moreover, selecting the correct computational procedure is also challenging for LLMs, as subtle differences in task descriptions or graph structures may require different algorithms. For example, numerical values associated with edges may represent either timestamps or weights, leading to entirely different graph computation tasks.
To this end, future benchmarks should evaluate LLM-based graph computation with implicit graph structures, bringing LLM-based graph computation closer to real-world scenarios.

\noindent {\em $\bullet$} \textit{Optimizing execution pipelines for complex graph queries.} Complex graph queries often require multi-step operations, whose execution order strongly affects computation cost and accuracy \cite{wei2026graphchain}. Refining task-relevant intermediate subgraphs before complex steps improves efficiency and precision \cite{lyu2025modular}. 
Without such optimization, executor-based methods suffer from inflated token consumption and diminished accuracy, while planner-based methods experience longer runtimes when large subgraphs are fed into high-complexity operations.
Therefore, future work should enable LLMs to decompose queries into optimized sub-task sequences, thereby enabling more scalable, efficient, and robust multi-step graph computation.


\noindent {\em $\bullet$} \textit{Privacy vulnerabilities.}
LLMs have been proven vulnerable to many attacks \cite{chen2025survey, yao2024survey, yu2025survey}, while real-world graphs capture sensitive user information or relationships \cite{imola2022differentially, zhang2025truss}.
For example, in e-commerce networks, vertices may include private attributes like home address, and edges reveal purchase or transaction records.
In healthcare systems, the graph not only stores patient data within its vertices, but also its edges expose the patient-doctor associations.
Through jailbreaking attacks, the graphs in the input prompt can be exposed to unauthorized users \cite{priyanshu2023chatbots, tang2023privacy}. In addition, membership inference attacks may reveal whether specific vertices or edges appear in the training corpus \cite{das2025security}.
It is essential to leverage LLMs for graph computation while ensuring data privacy.

\noindent {\em $\bullet$} \textit{LLMs for domain-specific graph computation.} 
Many real-world applications involve domain-specific graphs beyond conventional structures, with features such as rich attributes, heterogeneity, and temporal dependencies.
For example, in a global metabolic network \cite{peregrin2009conservation, barabasi2011network}, vertices encode enzymes, with colors denoting metabolic superclasses and sizes indicating associated genome counts.
In an attack graph \cite{ou2006scalable, kaynar2015distributed}, attributes of vertices indicate types, security conditions, vulnerabilities, and exploitability scores, while the direction of edges captures logical dependencies and attacker movements.
However, due to the limited generalizability of current LLMs \cite{lu2025fine}, directly applying LLMs designed for general graph computation to these highly specialized graphs often results in low accuracy and severe hallucinations. Enabling LLMs to handle richer graph structures could improve their utility in complex real-world scenarios.


\section{Conclusions}
\label{sec:conclusion}
We systematically reviewed the status quo of LLMs for graph computation. Specifically, we introduced a new taxonomy based on the operational roles of models and provided a thorough review and comparison of methods within each paradigm.
We summarized available datasets, source code, and experimental performance of existing methods. Our review suggested that while LLMs perform well on small-scale graphs and simple tasks, they still exhibit limitations when faced with large-scale or high-precision demands, indicating opportunities for future research and improvement.

\section*{Limitations}
As a survey paper, its limitations reflect the limitations of the LLM-based graph computation area. 
Although recent studies have made substantial progress in enabling LLMs to accurately solve graph computation tasks, LLMs still face substantial challenges when applied to graph-structured problems, including limited scalability, weak generalization, and unreliable reasoning over complex structures. In addition, many important graph problems have not been sufficiently covered by existing studies, such as community search and other advanced graph query tasks.


Besides, due to space constraints in the main text, several detailed summaries are placed in the appendices. These include representative graph computation tasks, benchmark datasets, and existing LLM-based graph computation methods. The task summary provides definitions of common graph computation problems and their computational complexities. The benchmark overview reports dataset statistics, including graph size, graph type, and task type. The method summary collects the source code availability of each method and summarizes the experimental results reported in the corresponding papers. We believe these materials are important and useful for readers.

\bibliography{custom}

\appendix

\section{Scope Definition and Paper Selection}
\label{sec:survey_comparison}

We collected candidate papers from major academic databases and preprint archives, including ACL Anthology, arXiv, Google Scholar, Semantic Scholar, DBLP, and OpenReview. The primary criterion for inclusion was an explicit focus on LLMs participating in algorithmic graph computational tasks. To maintain this focused scope, we distinguish our survey from the broader literature on traditional graph learning tasks (e.g., graph classification, link prediction, or node classification), as well as from purely multimodal approaches without direct LLM integration.








\begin{table*}[htbp]
    \centering
    \small
    \renewcommand{\arraystretch}{1.15}
    \caption{Summary of representative graph computation tasks, their definitions, and computational complexity. 
    }
    \label{tab:graph_tasks}
    \setlength{\tabcolsep}{2pt}
    \begin{tabular}{
        p{3.1cm}        
        p{10.8cm}       
        >{\centering\arraybackslash}p{2.0cm}     
    }
    \toprule
    \textbf{Task} & \textbf{Definition} & \textbf{Time Complexity} \\
    \midrule

        \multicolumn{3}{l}{\textit{Linear-time graph computation tasks}} \\
        \hline

        Vertex Degree 
        & Given an undirected graph $G$ and a vertex $v$, return $\deg(v)$.
        & $O(n)$ \\

        Edge Existence
        & Given a graph $G$ and vertices $u,v$, determine whether edge $(u,v)$ exists in $G$.
        & $O(n)$ \\

        Common Neighbors
        & Given a graph $G$ and vertices $u,v$, return their common neighbors.
        & $O(n)$ \\

        Graph Traversal
        & Given a graph $G$ and a source vertex $s$, visit every vertex from $s$.
        & $O(n+m)$ \\

        Connectivity
        & Given an undirected graph $G$ and vertices $u,v$, determine whether a path connects them.
        & $O(n+m)$ \\

        Connected Components
        & Given a graph $G$, partition its vertices into maximal connected components.
        & $O(n+m)$ \\


        Cycle Detection
        & Given a directed graph $G$, determine whether it contains a directed cycle.
        & $O(n+m)$ \\

        Bipartite Graph Check
        & Given a graph $G$, determine whether its vertices can be divided into two sets such that no edge connects vertices within the same set.
        & $O(n+m)$ \\

        Eulerian Path / Circuit
        & Given a graph $G$, determine whether it has a trail that visits every edge exactly once.
        & $O(n+m)$ \\

        Topological Sort
        & Given a directed acyclic $G$, output a linear ordering of all vertices in $V$ such that for every directed edge $(u,v)\in E$, vertex $u$ appears before vertex $v$ in the ordering.
        & $O(n+m)$ \\

        \hline
        \addlinespace[2pt]
        \multicolumn{3}{l}{\textit{Polynomial-time graph computation tasks}} \\
        \hline

        Shortest Path
        & Given an edge-weighted graph $G$,  source vertex $u$, and target vertex $v$, find a minimum-weight path from $u$ to $v$.
        & $O(m+n\log n)$ \\

        Graph Diameter
        & Given a graph $G$, compute the largest shortest-path distance between two vertices.
        & $O(n(n+m))$ \\

        Minimum Spanning Tree
        & Given an edge-weighted graph $G$, find a spanning tree that connects all vertices and has the minimum possible total edge weight.
        & $O(m)$ \\

        Maximum Triangle Sum
        & Given a vertex-weighted graph $G$, find a triangle with maximum total vertex weight.
        & $O(m^{3/2})$ \\

        Maximum Flow
        & Given a directed graph $G$ with edge capacities, a source vertex $s$, and a sink vertex $t$, compute the maximum $s$--$t$ flow.
        & $O(nm^2)$ \\

        Bipartite Matching
        & Given a bipartite graph $G$, output a set of edges such that no two chosen edges share the same endpoint.
        & $O(m\sqrt{n})$ \\

        Triangle Counting
        & Given a graph $G$, count its triangles. A triangle is a cycle of length $3$.
        & $O(m^{3/2})$ \\

        Clustering Coefficient
        & Given a graph $G$, compute exact local or global clustering coefficients that measure how likely the neighbors of a vertex are to be connected to each other.
        & $O(m^{3/2})$ \\

        PageRank
        & Given a directed graph $G$, compute PageRank scores for $T$ power iterations.
        & $O(T(n+m))$ \\

        \hline
        \addlinespace[2pt]
        \multicolumn{3}{l}{\textit{NP-complete graph computation tasks}} \\
        \hline

        Hamiltonian Path / Cycle
        & Given a graph $G$, determine whether there exists a simple path or cycle that visits every vertex in $G$ exactly once.
        & NP-complete \\

        Traveling Salesman Problem
        & Given a complete weighted graph $G$ and a bound $B$, determine whether $G$ has a Hamiltonian cycle of weight at most $B$.
        & NP-complete \\

        Maximum Clique
        & Given a graph $G$ and an integer $r$, determine whether $G$ has a clique of size at least $r$.
        & NP-complete \\

        Maximum Independent Set
        & Given a graph $G$, determine whether there exists a subset of vertices such that no two vertices in the subset are adjacent.
        & NP-complete \\

        Minimum Vertex Cover
        & Given a graph $G$ and an integer $r$, determine whether there exists a subset of vertices of size at most $r$ such that every edge in $G$ has at least one endpoint in the subset.
        & NP-complete \\

        Graph Coloring
        & Given a graph $G$ and an integer $c\ge 3$, determine whether $G$ has a proper $c$-coloring.
        & NP-complete \\

        Dominating Set
        & Given a graph $G$ and an integer $r$, determine whether there exists a set of at most $r$ vertices such that every vertex in $G$ is either in the set or adjacent to a vertex in the set.
        & NP-complete \\

        Subgraph Matching
        & Given a host graph $G$ and a pattern graph $H$, determine whether $H$ is isomorphic to a subgraph of $G$.
        & NP-complete \\

        Graph Edit Distance
        & Given graphs $G_1$, $G_2$, and a bound $B$, determine whether $G_1$ can be transformed into $G_2$ using at most $B$ edit operations.
        & NP-complete \\

        Maximum Common Subgraph
        & Given graphs $G_1$, $G_2$, and an integer $r$, determine whether they have a common subgraph with at least $r$ vertices.
        & NP-complete \\

        \bottomrule
    \end{tabular}

\footnotesize

\raggedright
\vspace{1em}
\hspace*{1em}
$T$: the number of PageRank iterations.
\end{table*}

\section{Summary of Graph Computation Tasks}
\label{sec:graph_tasks}
A broad range of graph computation tasks has been explored in LLM-based methods, ranging from simple queries such as checking whether an edge exists to complex problems such as finding a Hamiltonian path. To provide a systematic overview of the tasks used in existing studies, we summarize representative graph tasks in Table \ref{tab:graph_tasks}, including the definition and computational complexity.

Following prior work on graph computation benchmarks \cite{chen2024graphwiz}, we categorize these tasks by computational complexity into linear-time, polynomial-time, and NP-complete tasks. Because computational complexity captures the intrinsic algorithmic difficulty of a task, it directly reflects the different reasoning requirements imposed on LLMs. Consequently, organizing tasks by complexity provides a natural framework for comparing the reasoning abilities of LLMs in handling graph computation. 
In complexity analysis, we refer to standard graph algorithms and denote the number of vertices and edges by $n=|V|$ and $m=|E|$, respectively, and the degree of vertex $v$ is denoted as $\deg(v)$.

\noindent {\em -- } Linear-time graph tasks are tasks that can be solved in time proportional to the size of the graph, typically $O(n+m)$, or proportional to the size of a local neighborhood, such as $O(\deg(v))$. Solving these tasks requires LLMs to inspect vertices and edges or perform standard graph traversal.

\noindent {\em -- } Polynomial-time graph tasks are tasks that can be solved by algorithms whose running time is bounded by a polynomial function of the graph size, such as $O(n^2)$, $O(n^3)$, or $O(m\log n)$. These tasks usually require LLMs to execute multi-step logical deduction and track intermediate states.

\noindent {\em -- } NP-complete tasks represent a much higher level of computational difficulty, as no polynomial-time algorithms are known to exist for these problems. These tasks require LLMs to perform complex reasoning and often demand abilities such as dynamic backtracking and systematic exploration of possible solutions.
\newpage

\section{Introduction of Benchmark Datasets}
\label{sec:benchmark}

In this section, we list benchmark datasets for LLMs for graph computations.
Table \ref{tab:benchmark} summarizes representative benchmark datasets that have been used to evaluate large language models for graph computation.
The graph size column reports the approximate scale of graphs used in each dataset. 
The graph type column describes the structural or semantic properties of the graphs, such as whether they are directed, weighted, dynamic, heterogeneous, or attributed.
The graph source column reports whether the graphs are synthetic or derived from real-world data.
The meaning of the abbreviations used in the graph type and graph source columns is given in Table \ref{tab:konect_signs}.
The task coverage columns indicate whether the dataset includes linear-time, polynomial-time, or NP-complete graph computation tasks. Finally, the number of tasks column reports the total number of graph computation tasks included in each dataset when available.

\begin{table}[H]
    \centering
    \caption{Signs used in Table~\ref{tab:benchmark}.}
    \vspace{-2mm}
    \label{tab:konect_signs}
    \small
    \begin{tabular}{ll}
        \toprule
        \textbf{Sign} & \textbf{Meaning} \\
        \midrule

        \multicolumn{2}{l}{\textit{Network Format}} \\[2pt]
        U            & Unipartite, undirected, and unweighted graph \\
        D            & Directed graph \\
        D$_{\mathrm{acyc}}$ & Directed acyclic graph \\
        B            & Bipartite graph \\
        $+$          & Weighted or capacitated graph \\
        T            & Dynamic or temporal graph \\
        Het          & Heterogeneous graph \\
        Attr         & Attributed or text-attributed graph \\
        \addlinespace[4pt] 

        \multicolumn{2}{l}{\textit{Graph Source}} \\[2pt]
        RW           & Real-world graph \\
        Syn          & Synthetic graph \\
        \bottomrule
    \end{tabular}
\end{table}

\begin{table*}
    \centering
    \caption{Summary of benchmark datasets of large language models for graph computation.}
    \label{tab:benchmark}
    \setlength{\tabcolsep}{6pt} 
    \small
    \begin{tabular}{  
        p{6cm}       
        p{0.9cm}     
        p{2.3cm}     
        p{1.2cm}       
        >{\centering\arraybackslash}p{0.6cm}   
        >{\centering\arraybackslash}p{0.6cm}   
        >{\centering\arraybackslash}p{0.6cm}     
        >{\centering\arraybackslash}p{1cm}   
        }
        \toprule
        \multirow{2}{*}{ \textbf{Name}}
        & \multicolumn{3}{c}{\textbf{Graphs}}
        & \multicolumn{3}{c}{\textbf{Tasks}}
        &\multirow{2}{*}{ \makecell{\textbf{\# of}\\ \textbf{Tasks}}}\\
        \cmidrule(lr){2-4}
        \cmidrule(lr){5-7}
        & \textbf{Size}& \textbf{Type} & \textbf{Source}
        & \textbf{Linear} & \textbf{Poly} & \makebox[0.88cm][c]{\textbf{NP-Comp}}  & \\
        \midrule

        GPT4Graph \cite{guo2023gpt4graph}
        & $\sim 10^{1}$ & U, D, Attr & RW
        & \checkmark & \checkmark & -- & 10 \\

        NLGraph \cite{wang2023can}
        & $\sim 10^{2}$ & U, D, $+$, B, D$_{\mathrm{acyc}}$ & Syn
        & \checkmark & \checkmark & \checkmark & 8 \\

        LLMtoGraph Eval. \cite{liu2023evaluating}
        & $\sim 10^{2}$ & U & Syn
        & \checkmark & \checkmark & -- & 5 \\

        GraphQA \cite{fatemi2023talk}
        & $\sim 10^{1}$ & U & Syn
        & \checkmark & -- & -- & 7 \\

        GraphLLM Eval. \cite{chai2025graphllm}
        & $\sim 10^{2}$ & U, B, Attr & RW
        & -- & \checkmark & -- & 4 \\

        LLM4DyG \cite{zhang2024llm4dyg}
        & $\sim 10^{1}$ & T, U & RW, Syn
        & -- & \checkmark & -- & 9 \\

        GVLQA / GITA \cite{wei2024gita}
        & $\sim 10^{1}$ & U, D, $+$, B & RW, Syn
        & \checkmark & \checkmark & \checkmark & 7 \\

        GraphWiz \cite{chen2024graphwiz}
        & $\sim 10^{2}$ & U, D, $+$, B & Syn
        & \checkmark & \checkmark & \checkmark & 9 \\

        GraphInstruct \cite{luo2024graphinstruct}
        & $\sim 10^{2}$ & U, D, $+$, B & Syn
        & \checkmark & \checkmark & \checkmark & 21 \\

        InstructGraph \cite{wang2024instructgraph}
        & $\sim 10^{2}$ & U, D, $+$, Het, Attr & RW, Syn
        & \checkmark & \checkmark & \checkmark & 29 \\

        NLGIFT \cite{zhang2024can}
        & $\sim 10^{1}$ & U, D, $+$ & Syn
        & \checkmark & \checkmark & -- & 4 \\

        ProGraph \cite{li2024can}
        & $\sim 10^{6}$ & U, D, $+$, Attr & RW
        & \checkmark & \checkmark & -- & 3 \\

        GraphArena \cite{tang2024grapharena}
        & $\sim 10^{2}$ & U, $+$ & RW
        & -- & \checkmark & \checkmark & 10 \\

        GraCoRe \cite{yuan2025gracore}
        & $\sim 10^{1}$ & U, D, $+$, Het, Attr & RW, Syn
        & \checkmark & \checkmark & \checkmark & 19 \\

        Graph Pattern Comp. \cite{dai2025large}
        & $\sim 10^{2}$ & U, D, Attr & RW, Syn
        & -- & \checkmark & -- & 11 \\

        GCoder / GraphWild \cite{zhang2024gcoder}
        & $\sim 10^{2}$ & U, D, $+$, B & RW, Syn
        & \checkmark & \checkmark & \checkmark & 13 \\

        GraphSilo / GraphPRM \cite{peng2025rewarding}
        & $\sim 10^{2}$ & U, D, $+$ & Syn
        & \checkmark & \checkmark & \checkmark & 13 \\

        GTools \cite{wang2025graphtool}
        & $\sim 10^{2}$ & U, D, $+$ & Syn
        & \checkmark & \checkmark & -- & 20 \\

        G-REAL / MA-GTS \cite{yuan2025ma}
        & $\sim 10^{1}$ & U, $+$ & RW
        & \checkmark & \checkmark & \checkmark & 4 \\

        GraphPile \cite{zhang2025improving}
        & $\sim 10^{2}$ & U, D, $+$, B & RW, Syn
        & \checkmark & \checkmark & -- & 23 \\

        GraphAlgorithm \cite{hu2025rethinking}
        & $\sim 10^{3}$ & U, D, $+$ & RW
        & \checkmark & \checkmark & \checkmark & 239 \\

        GT Bench / GTA \cite{xu2025gta}
        & $\sim 10^{2}$ & U, D, $+$ & Syn
        & \checkmark & \checkmark & \checkmark & 44 \\

        GraphOmni \cite{xu2025graphomni}
        & $\sim 10^{1}$ & U, B & Syn
        & \checkmark & \checkmark & -- & 6 \\

        G1 / Erd\H{o}s \cite{guo2025g1}
        & $\sim 10^{2}$ & U, D, $+$ & RW
        & \checkmark & \checkmark & \checkmark & 50 \\

        VGCURE \cite{zhu2025benchmarking}
        & $\sim 10^{1}$ & U, D, Attr & RW, Syn
        & \checkmark & \checkmark & -- & 22 \\

        GraphSQA / GraphInsight \cite{cao2025graphinsight}
        & $\sim 10^{2}$ & U, D, $+$ & Syn
        & \checkmark & -- & -- & 20 \\

        LLMTM \cite{hao2026llmtm}
        & $\sim 10^{2}$ & T, U & RW, Syn
        & -- & \checkmark & -- & 6 \\

        GRAlgoBench \cite{zhang2026gralgobench}
        & $\sim 10^{2}$ & U, $+$ & RW
        & \checkmark & \checkmark & \checkmark & 9 \\

        ComplexGraph / GraphSkill \cite{wang2026graphskill}
        & $\sim 10^{4}$ & U, D, D$_{\mathrm{acyc}}$, $+$ & Syn
        & \checkmark & \checkmark & \checkmark & 23 \\

        Grena / GraphVista \cite{han2026graphvista}
        & $\sim 10^{3}$ & U, $+$ & Syn
        & \checkmark & \checkmark & \checkmark & 20 \\

        Graph4real / GraphCogent \cite{wang2026graphcogent}
        & $\sim 10^{3}$ & U, D, $+$, Attr & RW
        & \checkmark & \checkmark & -- & 21 \\

        Less-is-More Graph Tasks \cite{shirai2025less}
        & $\sim 10^{2}$ & D, Attr & Syn
        & \checkmark & -- & -- & 8 \\

        GraphThought Dataset \cite{huang2025graphthought}
        & $\sim 10^{2}$ & U, $+$, Attr & RW
        & \checkmark & \checkmark & \checkmark & 10 \\

        NPG-Muse Corpus \cite{wang2025graph}
        & $\sim 10^{1}$ & U, $+$, Attr & Syn
        & -- & -- & \checkmark & 3 \\

        TRF Preference Dataset \cite{wei2025harnessing}
        & $\sim 10^{1}$ & U, D, $+$, B, D$_{\mathrm{acyc}}$ & Syn
        & \checkmark & \checkmark & \checkmark & 7 \\

        GTR Preference Dataset \cite{wei2026dynamicgtr}
        & $\sim 10^{1}$ & U, D, $+$, B, D$_{\mathrm{acyc}}$ & Syn
        & \checkmark & \checkmark & \checkmark & 7 \\

        \bottomrule
    \end{tabular}
\end{table*}

\newpage

\section{Summary of the methods for LLMs-based graph computation}
\label{sec:methods}
Table \ref{tab:method_summary} provides a comprehensive comparison of all methods reviewed in this survey. 
For each method, we indicate its publication venue, provide a summary of the experimental performance reported in the original paper, and specify whether an official implementation is publicly available.
For the experimental performance, we report both the maximum graph scale and accuracy. 

\noindent {\em -- } This size column indicates the largest graph size used for graph computations as reported by each paper. The scale is calculated as $\mathrm{round}(\log_{10} n_{\max})$, where $n_{\max}$ is the largest reported vertex count in graph computation tasks.

\noindent {\em -- } The accuracy columns summarize the reported performance of each method on three categories of graph computation tasks. These task categories follow the complexity-based grouping introduced in Appendix \ref{sec:graph_tasks}. 
Since different studies employ varying benchmarks and models, their reported raw accuracy is not directly comparable. 
To address this, we adopt a star-based rating to provide an overview of the reported effectiveness for each task category.
Each $\star$ corresponds to $20\%$ accuracy, with higher star ratings representing superior performance within the experimental setting of the respective paper. 
The symbol “--” means that the paper does not report results for that task category.

\noindent {\em -- } The code column provides links to the official implementation when available. 

\begin{table*}[t]
    \centering
    \caption{Summary of methods for LLM-based graph computation.
    }
    \label{tab:method_summary}
    \small
    \begingroup
    \setlength{\tabcolsep}{6pt}
    \begin{tabular}{@{}
        >{\centering\arraybackslash}p{0.68cm}
        >{\raggedright\arraybackslash}p{5.8cm}
        >{\centering\arraybackslash}p{3cm}
        >{\centering\arraybackslash}p{0.68cm}
        >{\centering\arraybackslash}p{0.88cm}
        >{\centering\arraybackslash}p{0.88cm}
        >{\centering\arraybackslash}p{0.88cm}
        >{\centering\arraybackslash}p{1cm}
        @{}}
        \toprule
        \multirow{2}{*}{\textbf{Cat.}}
        & \multirow{2}{*}{\textbf{Method}}
        & \multirow{2}{*}{\textbf{Venue}}
        & \multirow{2}{*}{\textbf{Size}}
        & \multicolumn{3}{c}{\textbf{Accuracy}} 
        &\multirow{2}{*}{ \textbf{Code}} \\
        \cmidrule(lr){5-7}
         &  &  &  &  Linear & Poly &\makebox[0.88cm][c]{NP-Complete} &  \\
        \midrule
\multirow{8}{*}{\rotatebox[origin=c]{90}{Prompting}}
    & NLGraph \cite{wang2023can} & NeurIPS'23 & $10^{2}$ & $\star\star\star\star$ & $\star\star\star$ & $\star\star$ & \href{https://github.com/Arthur-Heng/NLGraph}{\checkmark} \\
    & GPT4Graph \cite{guo2023gpt4graph} & arXiv'23 & $10^{1}$ & $\star\star\star\star$ & $\star\star$ & -- & -- \\
    & LLMtoGraph \cite{liu2023evaluating} & arXiv'23 & $10^{2}$ & $\star\star\star\star$ & $\star\star\star$ & -- & \href{https://github.com/Ayame1006/LLMtoGraph}{\checkmark} \\
    & LLM4DyG \cite{zhang2024llm4dyg} & KDD'24 & $10^{1}$ & -- & $\star\star\star$ & -- & \href{https://github.com/wondergo2017/LLM4DyG}{\checkmark} \\
    & GraphArena \cite{tang2024grapharena} & ICLR'25 & $10^{2}$ & -- & $\star\star\star\star$ & $\star\star$ & \href{https://github.com/squareRoot3/GraphArena}{\checkmark} \\
    & GraCoRe \cite{yuan2025gracore} & COLING'25 & $10^{1}$ & $\star\star\star\star\star$ & $\star\star\star$ & $\star\star\star\star$ & \href{https://github.com/ZIKEYUAN/GraCoRe}{\checkmark} \\
    & GraphPattern \cite{dai2025large} & ICLR'25 & $10^{2}$ & -- & $\star\star\star$ & -- & \href{https://github.com/DDigimon/GraphPattern}{\checkmark} \\
    & LLMTM \cite{hao2026llmtm} & AAAI'26 & $10^{2}$ & -- & $\star\star\star\star\star$ & -- & \href{https://github.com/Wjerry5/LLMTM}{\checkmark} \\

\midrule

\multirow{10}{*}{\rotatebox[origin=c]{90}{Encoding}}
    & Talk Like a Graph \cite{fatemi2023talk} & ICLR'24 & $10^{1}$ & $\star\star\star$ & -- & -- & -- \\
    & GraphLLM \cite{chai2025graphllm} & arXiv'23 & $10^{2}$ & -- & $\star\star\star\star\star$ & -- & \href{https://github.com/mistyreed63849/Graph-LLM}{\checkmark} \\
    & Graph-Token \cite{perozzi2024let} & arXiv'24 & $10^{1}$ & $\star\star\star\star$ & $\star\star\star$ & -- & -- \\
    & GraphDO \cite{ge2025can} & ACL'25 & $10^{1}$ & $\star\star\star\star$ & $\star\star\star$ & $\star\star\star$ & \href{https://github.com/YuyaoGe/GraphDO}{\checkmark} \\
    & GraphOmni \cite{xu2025graphomni} & ICLR'26 & $10^{1}$ & $\star\star\star\star\star$ & $\star\star\star\star$ & -- & \href{https://github.com/GAI-Community/GraphOmni}{\checkmark} \\
    & GraphInsight \cite{cao2025graphinsight} & ACL'25 & $10^{2}$ & $\star\star\star\star\star$ & -- & -- & -- \\
    & Graph Lineariz. \cite{xypolopoulos2024graph} & arXiv'24 & $10^{2}$ & $\star\star\star$ & $\star\star$ & -- & -- \\
    & CL-OWL \cite{zangari2026colorful} & arXiv'26 & $10^{2}$ & $\star\star\star\star\star$ & $\star\star\star$ & -- & \href{https://github.com/angelozangari/CL-OWL}{\checkmark} \\
    & DynamicTRF \cite{wei2025harnessing} & arXiv'25 & $10^{1}$ & $\star\star\star\star\star$ & $\star\star\star\star\star$ & $\star\star\star\star\star$ & -- \\
    & DynamicGTR \cite{wei2026dynamicgtr} & CVPR'26 & $10^{1}$ & $\star\star\star\star\star$ & $\star\star\star\star\star$ & $\star\star\star\star\star$ & -- \\

\midrule

\multirow{12}{*}{\rotatebox[origin=c]{90}{\shortstack{Post-training}}}
    & GraphWiz \cite{chen2024graphwiz} & KDD'24 & $10^{2}$ & $\star\star\star\star$ & $\star\star\star$ & $\star\star\star\star$ & \href{https://github.com/nuochenpku/Graph-Reasoning-LLM}{\checkmark} \\
    & GITA \cite{wei2024gita} & NeurIPS'24 & $10^{1}$ & $\star\star\star\star$ & $\star\star\star$ & $\star\star$ & \href{https://github.com/WEIYanbin1999/GITA}{\checkmark} \\
    & InstructGraph \cite{wang2024instructgraph} & ACL Findings'24 & $10^{2}$ & $\star\star\star\star$ & $\star\star\star\star$ & $\star\star$ & \href{https://github.com/wjn1996/InstructGraph}{\checkmark} \\
    & GraphInstruct \cite{luo2024graphinstruct} & arXiv'24 & $10^{2}$ & $\star\star\star\star$ & $\star\star\star$ & $\star\star\star$ & \href{https://github.com/CGCL-codes/GraphInstruct}{\checkmark} \\
    & NLGift \cite{zhang2024can} & EMNLP Findings'24 & $10^{1}$ & $\star\star\star\star\star$ & $\star\star\star$ & -- & \href{https://github.com/MatthewYZhang/NLGift}{\checkmark} \\
    & HLM-G \cite{khurana2024hierarchical} & arXiv'24 & $10^{2}$ & $\star\star\star\star\star$ & $\star\star\star\star\star$ & -- & -- \\
    & GraphPRM \cite{peng2025rewarding} & KDD'25 & $10^{2}$ & $\star\star\star$ & $\star$ & $\star\star\star$ & \href{https://github.com/GKNL/GraphPRM}{\checkmark} \\
    & G1 \cite{guo2025g1} & arXiv'25 & $10^{2}$ & $\star\star\star\star$ & $\star\star$ & $\star\star\star\star$ & \href{https://github.com/PKU-ML/G1}{\checkmark} \\
    & Post-training Alignment \cite{zhang2025generalizable} & arXiv'25 & $10^{1}$ & $\star\star\star\star\star$ & $\star\star\star\star\star$ & -- & \href{https://anonymous.4open.science/r/Graph_RL-BF08/readme.md}{\checkmark} \\
    & GraphThought \cite{huang2025graphthought} & arXiv'25 & $10^{2}$ & $\star\star\star\star\star$ & $\star\star\star\star\star$ & $\star\star\star$ & \href{https://anonymous.4open.science/r/GraphThought-7CFE}{\checkmark} \\
    & Less is More \cite{shirai2025less} & arXiv'25 & $10^{2}$ & $\star\star\star\star\star$ & -- & -- & -- \\
    & NPG-Muse \cite{wang2025graph} & arXiv'25 & $10^{1}$ & -- & -- & $\star\star\star$ & \href{https://github.com/littlewyy/NPG-Muse}{\checkmark} \\

\midrule

\multirow{7}{*}{\rotatebox[origin=c]{90}{\shortstack{Code\\generation}}}
    & GraphArena \cite{tang2024grapharena} & ICLR'25 & $10^{2}$ & -- & $\star\star\star$ & $\star\star$ & \href{https://github.com/squareRoot3/GraphArena}{\checkmark} \\
    & GCoder \cite{zhang2024gcoder} & CIKM'25 & $10^{6}$ & $\star\star\star\star\star$ & $\star\star\star\star\star$ & $\star\star\star\star\star$ & \href{https://github.com/Bklight999/WWW25-GCoder/tree/master}{\checkmark} \\
    & GRRAF \cite{li2025grraf} & EMNLP Findings'25 & $10^{4}$ & $\star\star\star\star\star$ & $\star\star\star\star\star$ & $\star\star\star\star\star$ & \href{https://github.com/hanklee97121/GRRAF/tree/main}{\checkmark} \\
    & PIE \cite{gong2025pseudocode} & arXiv'25 & $10^{2}$ & -- & $\star\star\star\star\star$ & $\star\star\star\star\star$ & -- \\
    & GraphMind \cite{zhang2025improving} & COLM'25 & $10^{2}$ & $\star\star\star\star$ & $\star$ & -- & -- \\
    & Simple-RTC \cite{hu2025rethinking} & arXiv'25 & $10^{3}$ & $\star\star\star\star\star$ & $\star\star\star\star\star$ & $\star\star\star\star\star$ & -- \\
    & GraphSkill \cite{wang2026graphskill} & arXiv'26 & $10^{4}$ & $\star\star\star\star\star$ & $\star\star\star\star\star$ & $\star\star\star\star\star$ & \href{https://github.com/FairyFali/GraphSkill}{\checkmark} \\

\midrule

\multirow{6}{*}{\rotatebox[origin=c]{90}{\shortstack{Function\\calling}}}
    & Graph-ToolFormer \cite{zhang2023graph} & arXiv'23 & $10^{1}$ & $\star\star\star\star$ & $\star\star\star\star$ & $\star\star\star\star$ & \href{https://github.com/jwzhanggy/Graph_Toolformer}{\checkmark} \\
    & Graph-CoT \cite{jin2024graphcot} & ACL Findings'24 & $10^{8}$ & -- & -- & -- & \href{https://github.com/PeterGriffinJin/Graph-CoT}{\checkmark} \\
    & LLM4Graph \cite{li2024can} & NeurIPS'24 & $10^{6}$ & $\star\star\star$ & $\star\star\star$ & -- & \href{https://github.com/BUPT-GAMMA/ProGraph}{\checkmark} \\
    & Graph-Grounded LLMs \cite{gupta2025graph} & arXiv'25 & $10^{2}$ & $\star\star\star\star\star$ & $\star\star\star\star\star$ & $\star\star\star\star\star$ & -- \\
    & GraphChain \cite{wei2026graphchain} & NeurIPS'25 & $10^{5}$ & $\star\star\star\star\star$ & $\star\star\star\star\star$ & -- & \href{https://github.com/wuanjunruc/GraphChain}{\checkmark} \\
    & GraphTool-Instruction \cite{wang2025graphtool} & KDD'25 & $10^{2}$ & $\star\star\star\star\star$ & $\star\star\star\star\star$ & -- & \href{https://github.com/RongzhengWang/GraphTool-Instruction}{\checkmark} \\
    & EGL-SCA \cite{yuan2026eglsca} & arXiv'26 & $10^{1}$ & $\star\star\star\star\star$ & $\star\star\star\star\star$ & $\star\star\star\star\star$ & -- \\

\midrule

\multirow{7}{*}{\rotatebox[origin=c]{90}{\shortstack{Multi-agent\\collaboration}}}
    & GraphTeam \cite{li2024graphteam} & arXiv'24 & $10^{6}$ & $\star\star\star\star\star$ & $\star\star\star\star\star$ & $\star\star\star\star\star$ & \href{https://github.com/BUPT-GAMMA/GraphTeam}{\checkmark} \\
    & GraphAgent-Reasoner \cite{hu2024scalable} & arXiv'24 & $10^{3}$ & $\star\star\star\star\star$ & $\star\star\star\star\star$ & -- & -- \\
    & MA-GTS \cite{yuan2025ma} & EMNLP'25 & $10^{2}$ & $\star\star\star\star\star$ & $\star\star\star\star\star$ & $\star\star\star\star\star$ & \href{https://github.com/ZIKEYUAN/MA-GTS.git}{\checkmark} \\
    & GTA \cite{xu2025gta} & arXiv'25 & $10^{2}$ & $\star\star\star\star\star$ & $\star\star\star\star\star$ & $\star\star\star\star$ & -- \\
    & GraphDC \cite{li2026graphdc} & arXiv'26 & $10^{2}$ & $\star\star\star\star$ & $\star\star\star\star$ & -- & \href{https://anonymous.4open.science/r/GraphDCA-FBDF}{\checkmark} \\
    & GraphVista \cite{han2026graphvista} & arXiv'25 & $10^{3}$ & $\star\star\star\star\star$ & $\star\star\star$ & $\star\star\star$ & -- \\
    & GraphCogent \cite{wang2026graphcogent} & WWW'26 & $10^{4}$ & $\star\star\star\star\star$ & $\star\star\star\star\star$ & -- & -- \\

\bottomrule
\end{tabular}

\noindent
\vspace{0.3em}
\parbox{\textwidth}{
\footnotesize
\raggedright
\textbf{--}: Not reported / not evaluated \hspace{2em}
\textbf{\checkmark}: Code available \hspace{2em}
$\star$: Average accuracy of this task category $(1$ to $5)$\\
Size is $10^{\mathrm{round}(\log_{10} n_{\max})}$, with .5 rounded up. Here $n_{\max}$ denotes the largest reported graph-computation vertex count.}
\endgroup
\end{table*}

\end{document}